\documentclass[12pt]{iopart}

\usepackage{array, makecell}
\usepackage{algorithmic}
\usepackage{algorithm}
\usepackage{multirow}
\usepackage{graphicx}
\usepackage{caption}
\usepackage{textcomp}
\usepackage{booktabs}
\usepackage{tabularray}
\usepackage{url}
\usepackage{rotating}

\begin{document}

\title[Feature Selection via Robust Weighted Score...]{Feature Selection via Robust Weighted Score for High Dimensional Binary Class-Imbalanced Gene Expression Data}

\author{Zardad Khan$^{1*}$, Amjad Ali$^{1}$, Saeed Aldahmani$^{1}$}
\address{$^1$Department of Statistics and Business Analytics, United Arab Emirates University, Al Ain, UAE}

\ead{\textmd{Zardad Khan:} zaar@uaeu.ac.ae; \textmd{Saeed Aldahmani:} saldahmani@uaeu.ac.ae}

\vspace{10pt}
\begin{indented}
	\item[]June 2023
\end{indented}

\begin{abstract}
In this paper, a robust weighted score for unbalanced data (ROWSU) is proposed for selecting the most discriminative feature for high dimensional gene expression binary classification with class-imbalance problem. The method addresses one of the most challenging problems of highly skewed class distributions in gene expression datasets that adversely affect the performance of classification algorithms. First, the training dataset is balanced by synthetically generating data points from minority class observations. Second, a minimum subset of genes is selected using a greedy search approach. Third, a novel wieghted robust score, where the weights are computed by support vectors, is introduced to obtain a refined set of genes. The highest scoring genes based on this approach are combined with the minimum subset of genes selected by the greedy search approach to form the final set of genes. The novel method ensures the selection of the most discriminative genes, even in the presence of skewed class distribution, thus improving the performance of the classifiers. The performance of the proposed ROWSU method is evaluated on $6$ gene expression datasets. Classification accuracy and sensitivity are used as performance metrics to compare the proposed ROWSU algorithm with several other state-of-the-art methods. Boxplots and stability plots are also constructed for a better understanding of the results. The results show that the proposed method outperforms the existing feature selection procedures based on classification performance from $k$ nearest neighbours ($k$NN) and random forest (RF) classifiers.

\end{abstract}

\vspace{2pc}

%
%
%
%
%

\section{Introduction}
\label{introduction}
Gene expression datasets have an important role in enabling researchers to diagnose diseases and investigate the complexities of biological processes. These datasets, generated through deoxyribonucleic acid (DNA) microarrays and ribonucleic acid (RNA) sequencing, provide important insights into gene activities of various organism cells. They comprise a large number of features/genes, many of which are redundant and noisy, and may not play a significant role in diagnosing a specific disease or biological process. To overcome this issue, it is necessary to select a small set of genes that have the most discriminative power to classify the samples into their correct classes (i.e., the organisms from which biopsies/tissues are taken have the disease or not). Feature selection has several advantages, such as increasing accuracy and reducing the execution time of classification models. Additionally, feature selection can also reduce the curse of dimensionality and increase the generality of classifiers. There are three main categories of feature selection techniques given as follows:
\begin{enumerate}
\item \textbf{Embedded methods:} These procedures integrate feature selection within the training process of a machine learning classifier. Examples of the embedded methods are regularization techniques like least absolute shrinkage and selection operator (LASSO) \cite{tibshirani1996regression}, ridge regression \cite{hoerl1970ridge} and decision tree-based feature importance \cite{breiman1984classification}.

\item \textbf{Wrapper methods:} This class of feature selection methods computes all possible feature subsets using a specific machine learning algorithm and choose the features for which the fitted model gives high performance. Forward selection \cite{sutter1993comparison, blanchet2008forward}, backward elimination \cite{sutter1993comparison}, and recursive feature elimination \cite{chen2007enhanced} are the examples of the wrapper methods.

\item \textbf{Filter methods:} Filter methods select the most discriminative features by using statistical computations. These procedures determine the association of each feature with the response variable, which is used as a relevance score for the features. The Pearson product-moment correlation, Relief-based algorithms \cite{urbanowicz2018relief}, uncorrelated shrunken centroid (USC) \cite{yeung2003multiclass} and minimum redundancy-maximum relevance (MRMR) \cite{ding2005minimum} algorithms are the examples of filter methods.
\end{enumerate}

The above feature selection procedures have numerous advantages across various real-world applications. Selecting the most discriminative features not only simplifies the model but also enhances its performance and interpretability and allowing the researchers to obtain important insights. Moreover, feature selection plays an important role in preventing classifiers from overfitting and enables to generalize the models effectively for new data. However, these procedures fail to perform well when the data has an unbalanced class distribution. Class-imbalanced problem is very common in gene expression datasets, especially when dealing with exceptionally rare diseases \cite{bolon2014review}. The unbalanced class distribution can skew the features importance which leads to biased selection of features. In such situations, the unbalance problem may cause an over selection of majority class at the expense of minority class. To avoid the issue of skewed class distributions, there are two widely used remedies. The first remedy is balancing the data through oversampling the minority class observations or undersampling the majority class observations which generates a more equitable distribution \cite{feng2023novel, kamalov2023feature}. The second remedy is to use specialized techniques designed to avoid unbalanced problem such as cost-sensitive learning procedures or ensemble algorithms \cite{weiss2007cost, feng2020using, groccia2023cost, mohapatra2023application, nekouie2023new}. These strategies can enhance the models performance and ensure more accurate results in various unbalanced gene expression datasets.

This work proposes a filtering feature selection procedure for unbalanced gene expression datasets. The novel method named as robust weighted score for unbalanced dataset (ROWSU) selects the most relevant genes by using the following steps. In the first step, it balances the dataset through sub-sampling from the minority class to generate new data points by applying the ordinary mean function on the features in the samples. These data points are then added to the original data which makes it balanced. In the second step, a minimum subset of features is chosen through a greedy search approach given in \cite{mahmoud2014feature}. In the third step, a novel robust weighted score is used to find the final set of genes. The proposed robust weighted score is the hybridization of a novel robust Fisher type score with the corresponding weights computed through support vectors. The highest scoring genes are combined with the minimum subset of genes to get the final set of the most relevant features. The proposed method is designed to achieve an adequate classification performance when the datasets are extremely unbalanced. The performance of the ROWSU algorithm is assessed via $6$ benchmark problems. Classification accuracy and sensitivity are used to evaluate the performance of the proposed method in comparison with ordinary procedures, i.e., Fisher score (Fish) \cite{duda2001pattern, gu2012generalized}, Wilcoxon rank sum test (Wilc) \cite{lausen2004assessment,liao2006gene}, signal to noise ratio (SNR) \cite{mishra2011feature}, proportion overlapping score (POS) \cite{mahmoud2014feature} and maximum relevance-minimum redundancy (MRMR) \cite{ding2005minimum}. Stability plots and boxplots of the results are also constructed for a visual comparison of the models' performance. The results demonstrate the efficacy of the new algorithm which outperforms the other methods in the majority of the cases while using random forest (RF) \cite{breiman2001random} and $k$ nearest neighbours ($k$NN) \cite{cover1967nearest} for classification purpose.

The remainder of the manuscript is organized as follows: Section \ref{literature} summarizes the related work and Section \ref{methods} gives details of the proposed algorithm. In Section \ref{results}, the experimental design and results of the proposed method and other state-of-the-art procedures are discussed. This section also provides a short summary of the considered datasets. Finally, the manuscript is concluded in Section \ref{conclusion}.

\section{Related work}
\label{literature}
Several methods have been proposed in the literature for feature selection in high dimensional gene expression datasets. Feature selection aims at identifying those genes which have the most discriminative power and increase the performance of the models \cite{guyon2003introduction}. Feature selection has a crucial impact on the analysis and reduces the execution time of the models \cite{xue2018nonlinear,chaudhari2018improving}. A relative importance procedure has been proposed in \cite{draminski2008monte}, which randomly selects subsets of features from the entire space and grows a large number of trees on each of the subsets. The constructed trees are assessed using a validation set of observations. Finally, features are selected based on which the trees correctly classify the maximum number of sample points into their correct classes. A two-stage grey wolf optimization procedure has been proposed for selecting the most relevant features in high-dimensional data problems \cite{shen2022two}. This procedure ensures classification performance with a small set of features and reduces training time of the models. Another method, known as minimum redundancy-maximum relevance (MRMR) has been proposed in the literature \cite{ding2005minimum} to determine the most discriminative features. This procedure attains maximum relevance with the response variable while minimizing redundancy. A minimum redundancy-maximum relevance ensemble (MRMRE) as an extension of the MRMR method is given in \cite{de2013mrmre}, and has shown improvements in results. Another feature selection procedure based on principal component analysis technique has been proposed in \cite{lu2011principal}. This method selects those features that are more discriminative with minimum component variation. A similar method, is given in \cite{talloen2007ni}, which uses factor analysis technique instead of the principal component analysis. Statistical tests like t-test and Wilcoxon rank-sum test have also been used to select the most relevant features \cite{lausen2004assessment, altman1994dangers}. The authors in \cite{apiletti2012maskedpainter} presented a feature selection method which computes gene mask for each features in the data by using the range of the core interval of gene expressions, which ensures the selection of the more relevant genes in classifying the data points to their correct classes, thus, avoiding ambiguity. This method selects a minimum subset of features that unambiguously classify the majority of the training observations to their classes correctly by using the gene masks and overlapping scores via the set covering technique. The minimum subset of features and genes having smallest overlapping scores are combined in the ultimate set of discriminative features. An extension of \cite{apiletti2012maskedpainter} has been made in \cite{mahmoud2014feature}, where the core interval of gene expression is computed by using a robust form of dispersion, i.e., interquartile range. The most relevant genes are identified through the proportion of overlapping observations in each class. Features with smaller POS scores are considered more discriminative and relevant to the target variable. Moreover,  the relative dominant classes for all features have also been identified. This associates each of the features with the class for which it has a powerful discriminative capability. Then, a set of the most relevant features is identified through the integration of the minimum subset of feature and POS score giving the final set of genes. Another study, as outlined in \cite{shaikh2023filter}, contributes to the evolving landscape of multi-label classification, building upon diverse feature selection methodologies and emphasizing the efficacy of specific techniques for improved classification outcomes. For further studies about feature selection, additional resources are available in the literature, such as \cite{Hanczár_2023, liu2021fast, motieghader2020mrna, cersonsky2021improving}. These methods have several real-life applications and demonstrate promising results. However, they face issues when dealing with class-imbalance in the data. 

Researchers have proposed numerous procedures to address class unbalance problems. These include balancing the data through resampling techniques and then applying feature selection methods \cite{haixiang2017learning}. These procedures are shown to significantly enhance the performance of existing procedures. In \cite{du2015feature}, the authors introduced a feature selection procedure for imbalanced data based on genetic search. While genetic search methods are known for their effectiveness in achieving good performance, it is important to note that they come with a substantial computational burden. The authors in \cite{yin2013feature} have proposed a feature selection technique for unbalanced data by using class decomposition. This procedure divides the majority class into smaller sub-classes, then uses the proposed technique to select the discriminative features. A method in \cite{maldonado2014feature}, introduces a wrapper algorithm addressing class imbalance using a balanced loss function. Another technique in \cite{yang2013ensemble} proposes an ensemble wrapper procedures with resampling for data balancing. \cite{yijing2016adapted} proposes an adaptive algorithm with BPSO and FCBF for feature selection, including resampling and ensemble classification. Moreover, a method presented in \cite{kamalov2018sensitivity}, where orthogonal variance decomposition is utilized for feature evaluation, specifically considering feature interactions under defined conditions. Other methods related to class-imbalance problems can be seen in \cite{liu2019classification, zhang2023empirical}. 

Considering the above discussion, we have proposed a feature selection procedure for unbalanced gene expression datasets. The novel method balances the data in the first step and then finds a minimum subset of features by using a greedy search approach in conjucntion with a novel robust weighted score. The proposed method thus tries to resolve the issue of class-imbalance while correctly classifying patients to their correct classes based on a small number of genes where the ordinary methods perform poorly in skewed class distribution scenarios.

\section{The Robust Weighted Score for Unbalanced Data (ROWSU)}
\label{methods}
Let $\mathcal{L}=(X, Y)$ be a given gene expression dataset where $X$ represents the feature matrix with a total of $n$ observations and $p$ genes/features, (i.e., $X = [e_{ij}]_{n \times p} \in \Re^{n \times p}$, where, $i = 1, 2, \ldots, n$ and $j = 1, 2, \ldots, p$), and $Y$ symbolizes the corresponding response with two-classes, i.e., $Y \in (-, +)$. The total number of sample points ($n$) comprises $n^-$ negative and $n^+$ positive classes, where the class-distribution is extremely unbalanced, i.e., $n^- >>> n^+$. To select the most discriminative genes, the following steps will be taken into consideration.

\subsection{Balancing The Data}
\label{balance}
To balance the given dataset $\mathcal{L}$, $m = n^- - n^+$ random sub-samples are selected from the minority class (+) each of size $n^\prime$, i.e., $X_s$, where $s = 1, 2, \ldots, m$, since there are $p$ features and $n^\prime$ observations in each of the sub-samples. Therefore, the mean of the $j^{th}$ gene in the $s^{th}$ sample can be calculated as follows:
$$\bar{G}_{sj} = \frac{\sum_{i=1}^{n^\prime} e_{ij}}{n^\prime},$$
where $j = 1, 2, \ldots, p$ and $s = 1, 2, \ldots, m$. In this way, new observations are generated to increase the size of the minority class (+), i.e., 
$$\bar{X}_s = [\bar{G}_{s1}, \bar{G}_{s2}, \ldots, \bar{G}_{sp}]_{1 \times p},$$
where $s = 1, 2, \ldots, m$. The generated data points can be written in a matrix form as follows: 

\[
\hspace{4cm}
\bar{X} = \left[
\begin{array}{c}
	\bar{X}_1 \\
	\bar{X}_2 \\
	\vdots\\
	\bar{X}_m 
\end{array}
\right]
,\]
\[
\hspace{2cm}
 \Rightarrow  \bar{X} = \left[
\begin{array}{ccccc}
	\bar{G}_{1 \times 1} & \bar{G}_{1 \times 2} & \ldots& \bar{G}_{1 \times p} \\
	\bar{G}_{2 \times 1} & \bar{G}_{2 \times 2} & \ldots & \bar{G}_{2 \times p} \\
	\vdots & \vdots & \ddots & \vdots \\ 
	\bar{G}_{m \times 1} & \bar{G}_{m \times 2} & \ldots & \bar{G}_{m \times p}  
\end{array}
\right]
.\]
Now, write the generated data points with their class label ($+$), i.e., $$\mathcal{L^\prime} = (\bar{X}, Y=+).$$ 
Combine the generated data $\mathcal{L^\prime}$ and the given training data $\mathcal{L}$, i.e., $$\mathcal{L^*} = \mathcal{L} \cup \mathcal{L^\prime}.$$
Data $\mathcal{L^*}$ is balanced, consisting of a total of $n^*$ sample points, where each class has the same number of observations, i.e., 
 $$n^* = n + m,$$
 $$\Rightarrow n^* = (n^- + n^+) + (n^- - n^+),$$
 $$\Rightarrow n^* = 2n^-.$$
In $\mathcal{L^*}$, there are $n^-$ data points of each class. Use $\mathcal{L^*}$ instead of the original data $\mathcal{L}$ to select the most discriminative features.

\subsection{Selecting Minimum Subset of Features}
\label{mini}
To select a minimum subset of $p^\prime$ features from the balanced data $\mathcal{L^*}$ given in Subsection \ref{balance}, this work has used the greedy search approach given in \cite{mahmoud2014feature}. The selected subset of $p^\prime$ features must accurately classify the majority of the data points in the training dataset while averting the impact of outliers in gene expression values. The selection process uses gene masks and proportion overlapping scores (POS) \cite{mahmoud2014feature}. Initially, the gene with the maximum number of 1-bits in its mask is included in the subset. If there are multiple genes with the same number of 1-bits, the gene with the smallest POS score is added to the subset. Subsequently, the logical operator $AND$ updates the gene masks of the remaining genes to select the next one. Repeat this process until a specified number of genes is reached or no gene has 1-bit in its gene mask. For further discussion on the greedy search approach, refer to \cite{mahmoud2014feature}. For computational convenience, some notations are provided as follows:

$\mathcal{L^*}_{min}$: Dataset having minimum subset of $p^\prime$ features only.

$\mathcal{L^*}_{Rem}$: Dataset without minimum subset of genes, i.e., having $p^{\prime\prime} = p - p^\prime$ features.

\subsection{Robust Fisher Score}
In this work a robust Fisher (RFish) score is proposed for binary class problems (standard Fisher score can be seen in \cite{duda2001pattern} and \cite{gu2012generalized}). The proposed RFish score is given in the following equation:
\begin{equation}
	\psi_{j}= \frac{n^+|\tilde{\mu}_j^+ -\tilde{\mu}_j|+n^-|\tilde{\mu}_j^- -\tilde{\mu}_j|}{n^+ \Delta_j^+ + n^- \Delta_j^-}, \hspace{0.3cm}\textmd{where,} \hspace{0.3cm} j = 1, 2, \ldots, p^{\prime\prime},
	\label{rfish1}
\end{equation}
where $\psi_{j}$ shows robust fisher score for $j^{th}$ feature, $\tilde{\mu}_j^+$ and $\tilde{\mu}_j^-$ are the medians of $j^{th}$ gene for class $+$ and $-$, respectively. $\tilde{\mu}_j$ is the global median for $j^{th}$ gene in the given data. Furthermore, $\Delta_j^+$ and $\Delta_j^-$ are the mean absolute deviations of $j^{th}$ gene for class $+$ and $-$, respectively. 

Since, the data $\mathcal{L^*}$ has a balanced class distribution, i.e., both classes have $n^-$ sample points. Therefore, Equation \ref{rfish1} can be written as follows:
\begin{equation}
	\psi_{j}= \frac{|\tilde{\mu}_j^+ -\tilde{\mu}_j|+|\tilde{\mu}_j^- -\tilde{\mu}_j|}{ \Delta_j^+ + \Delta_j^-}, \hspace{0.3cm}\textmd{where,} \hspace{0.3cm} j = 1, 2, \ldots, p^{\prime\prime},
	\label{rfish2}
\end{equation}
\begin{equation}
	\Rightarrow \psi= \{\psi_{1}, \psi_{2}, \ldots, \psi_{p^{\prime\prime}}\}.
	\label{rfish3}
\end{equation}

\subsection{Feature Weights ($w$)}
For weighting the above score to increase its discriminating ability, the notion of support vectors is exploited \cite{hearst1998support,chang2011libsvm}. This method uses support vectors to determine an optimal hyperplane that classifies the observations into their correct classes. The hyperplane can be expressed as follows:
\begin{equation}
    \mathcal{H}: w^T.(\mathcal{M}(x_i))+b,
	\label{svm1}
\end{equation}
where $w$ shows the normal vector to the hyperplane, $x_i$ stands for the data point and $b$ represents the bias term, while $\mathcal{M}(x_i)$ maps $x_i$ into a higher dimensional space. 

The distance between the hyperplane $\mathcal{H}$ provided in Equation \ref{svm1} and a specific sample point $x_i$ is given by the following formula:
\begin{equation}
	\delta_\mathcal{H}(\mathcal{M}(x_i))=\frac{\left|w^T.(\mathcal{M}(x_i))+b \right|}{||w||_2},
	\label{svm2}
\end{equation}
where $||w||_2$ shows Euclidean norm expressed as:
\begin{equation*}
||w||_2 = \sqrt{w_1^2 + w_2^2 + \ldots + w_p^2}.
	\label{svm3}
\end{equation*}
The weight vector $w$ of features is determined by solving an optimization problem which maximize the distance $\delta_\mathcal{H}(x_i)$. The objective function is given by: 
$$\min \frac{1}{2}||w||_2, \hspace{0.2cm}\textmd{subject to the constraint} \hspace{0.2cm} y_i(w^T.\mathcal{M}(x_i)+b) \ge 1.$$ 
For the above optimization problem, the weight and $w$ is computed as
\begin{equation*}
	w = \sum_{i=1}^{n^*}\alpha_i y_i x_i, \textmd{     where $\alpha_i$ are Lagrange’s multipliers,}
\end{equation*}
\begin{equation}
	\Rightarrow w = \{w_1, w_2, \ldots, w_{p^{\prime\prime}}\}.
	\label{svm5}
\end{equation}

\subsection{The Robust Weighted Score}
\label{row}
The robust weighted score ($\phi_j$) for the $j^{th}$ feature is obtained by multiplying the RFish scores given in Equations \ref{rfish3} and feature weights computed in Equation \ref{svm5}, i.e., 
\begin{equation*}
	\phi_j = |w_j . \psi_j|  \hspace{0.3cm}\textmd{where} \hspace{0.3cm} j = 1, 2, \ldots, p^{\prime\prime},
	\label{final1}
\end{equation*}
\begin{equation}
	\Rightarrow \phi = \{\phi_1, \phi_2, \ldots, \phi_{p^{\prime\prime}}\}. 
	\label{final2}
\end{equation}
Arrange the weights $\phi$ computed in Equation \ref{final2} in descending order and select the $p^{\prime *}$ features with the highest score. Denote the data with the selected features by $\mathcal{L^*}_{Row}$.

\subsection{Final Set of the Most Discriminative Features}
Combine the minimum subset of genes selected by the greedy search approach in Subsection \ref{mini} and features selected by the Robust Weighted Score in Subsection \ref{row}, i.e., $p^{*} = p^{\prime} + p^{\prime *}$ and the final data becomes:
$$\mathcal{L^*}_{Final} = \mathcal{L^*}_{Min} \cup \mathcal{L^*}_{Row}.$$
$\mathcal{L^*}_{Final}$ will be used for model fitting to predict unseen data points.

\subsection{Algorithm}
\label{sec:Algorithm}
The steps taken by the proposed ROWSU algorithm are given as follows:
\begin{enumerate}
 \item Consider a training dataset with an unbalanced class distribution.
 \item Make the data balanced by using the technique proposed in Subsection \ref{balance}.
 \item Select the most discriminative minimum subset of $p^\prime$ features by the greedy search approach given in \cite{mahmoud2014feature}.
 \item Select $p^{\prime *}$ features from the remaining data $\mathcal{L}_{Rem}^*$ by using the proposed robust weighted score presented in Subsection \ref{row}.
 \item Combine the selected features in Steps 3 and 4 to get the final data $\mathcal{L}_{Final}^*$ consisting of the most discriminative features.
\end{enumerate}

\begin{figure}
	\centering
	\fbox{\includegraphics[width=1\textwidth]{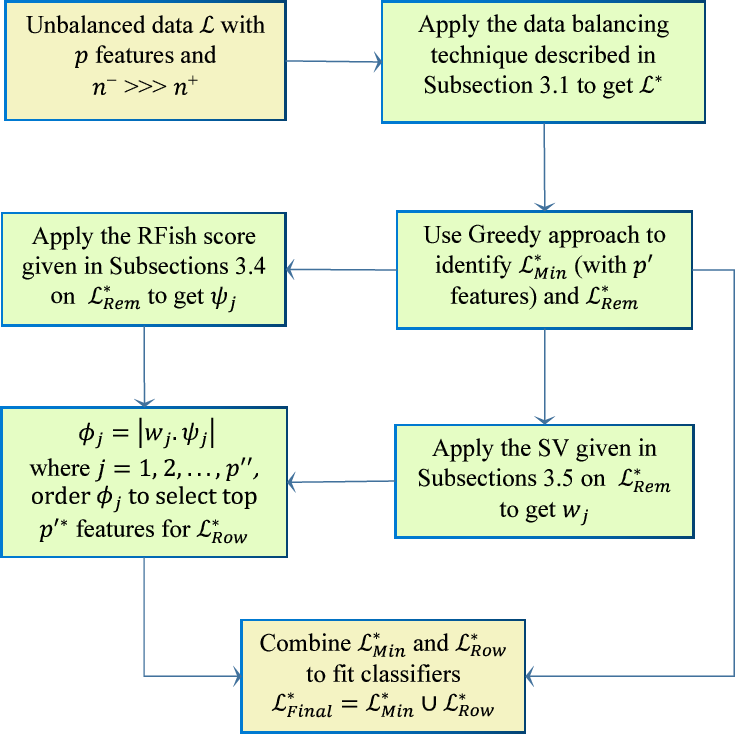}}
	\caption{Flowchart of the proposed ROWSU algorithm.}
	\label{Flowchart}
\end{figure}

Pseudo-code of the novel ROWSU method is provided in Algorithm \ref{Psudue}, and its flow chart in Figure \ref{Flowchart}.

\begin{algorithm}
	\caption{Pseudo code of the novel  ROWSU procedure.}
	\begin{algorithmic}[1]
		\STATE $\mathcal{L}$ $\leftarrow$ Training data with $p$ features and $n^- >>> n^+$ observations;
		
		\FOR {$s\gets 1:m$}
		\STATE $ X_s$ $\leftarrow$ Sub-sample of size $n^\prime$ from class $+$;
		\STATE $\bar{X}_s$ $\leftarrow$ Find column means of $ X_s$;
		\ENDFOR
		
		\STATE $\bar{X}$ $\leftarrow$ Merge the generated data points in a matrix form;
		\STATE $\mathcal{L}^*$ $\leftarrow$ Combine the original data and the generated data points;
		\STATE $\mathcal{L}_{Min}^*$ $\leftarrow$ Data consists of a minimum subset of $p^\prime$ genes selected by POS;
		\STATE $\mathcal{L}_{Rem}^*$ $\leftarrow$ Data without minimum subset;

		\FOR {$j\gets 1:p^{\prime\prime}$}
        \STATE $\psi_j$ $\leftarrow$ Compute robust Fisher score for $j^{th}$ gene in $\mathcal{L}_{Rem}^*$;
        \STATE $\psi$ $\leftarrow$ Save $\psi_j$ in a vector;
        \ENDFOR
        \STATE $\psi = \{\psi_1, \psi_2, \ldots, \psi_{p^{\prime\prime}}\}$ $\leftarrow$ Robust Fisher scores;
        
        \ /$*$ Using support vectors to compute feature weights $w$$ *$/
        \STATE $w = \{w_1, w_2, \ldots, w_{p^{\prime\prime}}\}$;
        
        /$*$ Robust weighted score $*$/
        \STATE $\phi_j = |w_j . \psi_j|$ where $j = 1, 2, \ldots, p^{\prime\prime}$ or $\phi = \{\phi_1, \phi_2, \ldots, \phi_{p^{\prime\prime}}\}$;
		\STATE Arrange in descending order and select the top ranked $p^{\prime *}$ genes;
		\STATE $p^{*} = p^{\prime} + p^{\prime *}$ $\leftarrow$ Final number of top ranked genes to be included in $\mathcal{L^*}_{Row}$;
		\STATE $\mathcal{L^*}_{Final} = \mathcal{L^*}_{Min} \cup \mathcal{L^*}_{Row}$ $\leftarrow$ The final data to be used for model construction.
	\end{algorithmic}
	\label{Psudue}
\end{algorithm}

\section{Experiments and results}
\label{results}
This section presents the experimental design, analyzed benchmark datasets, evaluation metrics and results of the proposed study. The results are demonstrated in a planned manner, which clarifies the outcomes of the analysis. Further details about the experimental setup and findings of the study are given in the coming subsections.

\subsection{Benchmark datasets}
\label{Benchmarkdatasets}
To evaluate the the performance of the proposed method and other existing procedures, 6 benchmark gene expression datasets are used. The considered datasets are briefly summarized in Table \ref{datasets}. The First column of the table represents data ID, the second column shows name of the data, while the third and fourth columns exhibits the number of observations $n$ and the number of features $p$, respectively. Class-wise distribution ($-, +$) is given in the fifth column, while the final column provides the data source.

\begin{table}
	\centering
	\caption{Benchmark datasets.}
	\fontsize{9.5}{9.5}\selectfont
	\renewcommand{\arraystretch}{1.2}
	\begin{tabular}{llcccl}
		\toprule
		ID & Data & $p$    & $n$  & ($-,+$)   &  Sources \\
		\midrule
        $D_1$ & DSRBCT    & 2308   & 63   & (43, 20)  & \url{https://tibshirani.su.domains/PAM/Rdist/khan.txt} \\
       $D_2$ & DLBCL     & 5469   & 77   & (58, 19)  & \url{https://www.openml.org/search?type=data&status=active&id=45088} \\
       $D_3$ & Leukemia  & 7129   & 72   & (47, 25)  & \url{https://www.openml.org/search?type=data&status=active&id=1104} \\
       $D_4$ & APEP      & 10936  & 130  & (69, 61)  & \url{https://www.openml.org/search?type=data&status=active&id=1141} \\
       $D_5$ & APOU      & 10936  & 201  & (124, 77) & \url{https://www.openml.org/search?type=data&status=active&id=1124} \\
       $D_6$ & Breast    & 4948   & 78   & (34, 44)  & \cite{michiels2005prediction} \\
		\bottomrule
	\end{tabular}%
	\label{datasets}%
\end{table}%

\subsection{Experimental setup}
\label{sec:Experimental setup}
The experimental design of the study is described in this section. For models assessment, 6 benchmark gene expression datasets have been analyzed. As the proposed method is designed for unbalanced gene expression datasets, therefore the observations are used in a 4:1 ratio, i.e., 80\% of negative ($-$) class and 20\% of positive ($+$) class. This is done by discarding minority class observations randomly, if the imbalanced ratio was not 4:1 in the original data. Each dataset is randomly divided into two groups, i.e., 80\% training and 20\% testing. The training part is used for model fitting, while evaluation is carried out on the testing part of the data. This splitting process is repeated 500 times for feature selection and classifiers construction. Fisher score (Fish), Wilcoxon rank sum test (Wilc), signal to noise ratio (SNR), proportion overlapping (POS) and maximum relevancy-minimum redundancy (MRMR) are used for comparison purposes. Moreover, random forest (RF) and $k$ nearest neighbours ($k$NN) have been used as classifiers based on the selected genes. Classification accuracy and sensitivity are the considered evaluation metrics. R programming is used for the experiments.

Furthermore, different numbers of the most discriminative features have been selected through the proposed method and the other competitors (i.e., $p^\prime = 5, 10, 15, 20, 25, 30$). These sets are used for fitting the considered classifiers to evaluate the performance of the proposed study.

\subsection{Results}
\label{sec:Results}
This section presents the results obtained from different feature selection algorithms, including the proposed ROWSU, Fish, Wilc, SNR, POS, and MRMR, along with the two classifiers random forest (RF) and $k$ nearest neighbors (kNN) on the $D_1$, $D_2$, $D_3$, $D_4$, $D_5$ and $D_6$ datasets. The performance metrics, accuracy and sensitivity are computed for various values of the selected features $p^*$.

Table \ref{d1} shows the results, computed for $D_1$ data for different numbers of features ($p^*$). The novel ROWSU procedure consistently outperforms the other feature selection methods. It is evident from the table that for RF, the novel method outperforms all the other procedures with classification accuracy ranging from 94.3\% to 97.2\%, and sensitivity from 77.6\% to 87.1\%. Specifically, ROWSU attains highest accuracy of 97.2\% and sensitivity of 87.1\% at $p^* = 20$. This demonstrates the effectiveness of the proposed method. The Fish procedure in RF classification attains about similar accuracy and sensitivity for some values of $p^*$, yet it fails to outperform the proposed method. In case of $k$NN, ROWSU method outperforms the other techniques with accuracy ranging from 97.2\% to 99.5\% and sensitivity from 89\% to 98.6\%. ROWSU attains its maximum accuracy of 98.6\% and sensitivity of 98.6\% at $p^* = 20$. The Fish method also gives comparable performance on RF, yet it falls slightly behind the proposed procedure. The rest of the procedures did not perform well on any of the classifiers.

Table \ref{d2} gives results for the $D_2$ dataset computed from the $k$NN and RF classifiers. In the case of RF, ROWSU gives remarkable accuracy ranging from 80.4\% to 89.2\% and sensitivity from 42.2\% to 72.3\%, as compared to the other methods. The proposed method consistently outperforms the other methods, demonstrating its potential to select features that result in correct estimation and accurately classifying the positive instances, which is highly desired in gene expression classification problems.  In the same way, the proposed method gives promising results in the case of $k$NN classifier among all the classical feature selection procedures. Accuracy of the proposed method ranges from 81.6\% to 89\% and sensitivity from 72.1\% to 92.8\%. Moreover, MRMR demonstrates high accuracy in for RF classifier, while Fish in the case of $k$NN classifier for  $p^* = 5$. The remaining methods did not provide satisfactory results in any of the cases.

Table \ref{d3} demonstrates the results computed for the $D_3$ dataset. In terms of accuracy, the novel ROWSU method consistently outperforms other techniques for various number of features ($p^*$) for both classifiers. ROWSU shows promising values of accuracy, ranging from 92.1\% to 93.9\% for RF and 92.4\% to 94.1\% for $k$NN, with maximum values at $p^* = 15$. In terms of sensitivity, the proposed ROWSU gives better performance than the others. It gives the highest sensitivity ranging from 76.3\% to 81.2\% for RF and 73.4\% to 83.1\% for kNN. POS gives maximum accuracy and sensitivity only at $p^* = 5$ for $k$NN. Moreover, Fish outperforms the others in terms of accuracy computed via RF at $p^* = 10$.

Table \ref{d4} demonstrates a detailed comparison of the novel method selecting various number ($p^*$) of features in comparison with the existing procedures on the $D_4$ dataset. The ROWSU method gives comparable performance to the MRMR in terms of accuracy in both RF and $k$NN classifications. In terms of sensitivity, the proposed method gives comparable performance to the existing procedures, i.e., Fish, POS and MRMR. It is evident from the table that the proposed method outperforms the classical procedures in the majority of the situations. It consistently shows competitive performance for different $p^*$ values.

\begin{figure}
	\centering
	\includegraphics[width=0.9\textwidth]{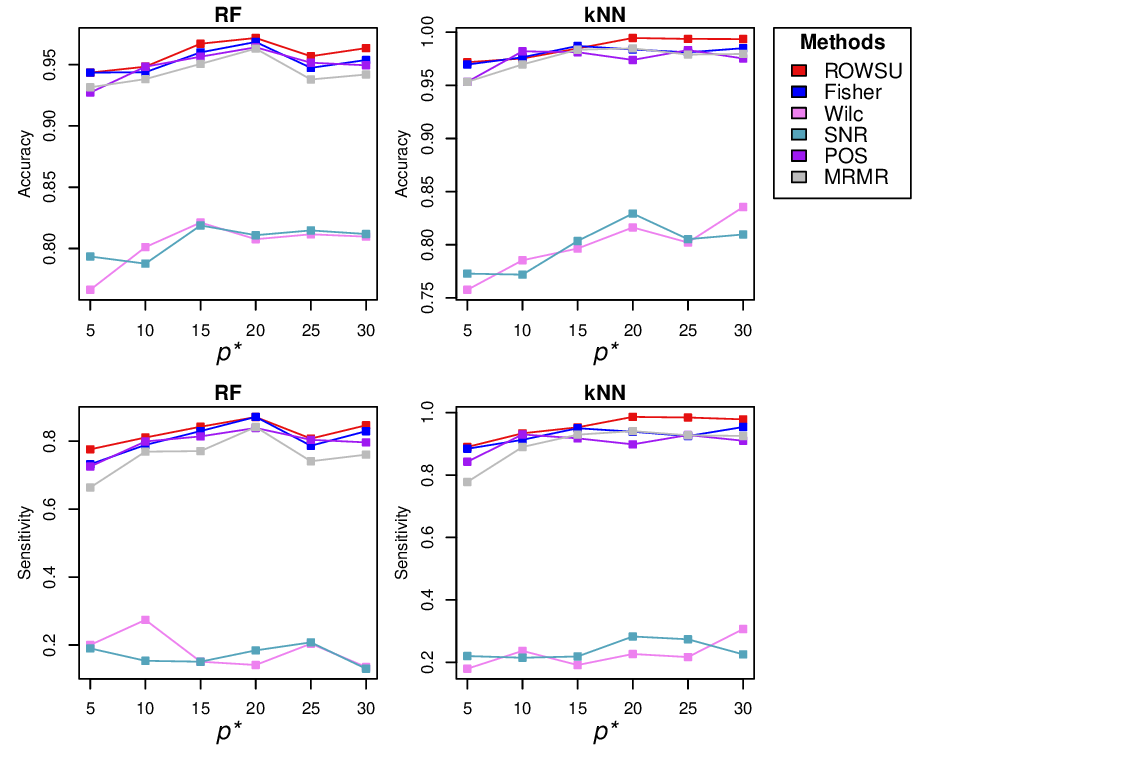}
	\caption{Matplots of the results computed for the top $p^*$ features by the proposed method and other state-of-the-art procedures using the \textbf{$D_1$} dataset.}
	\label{m1}
\end{figure}

\begin{figure}
    \centering
	\includegraphics[width=0.9\textwidth]{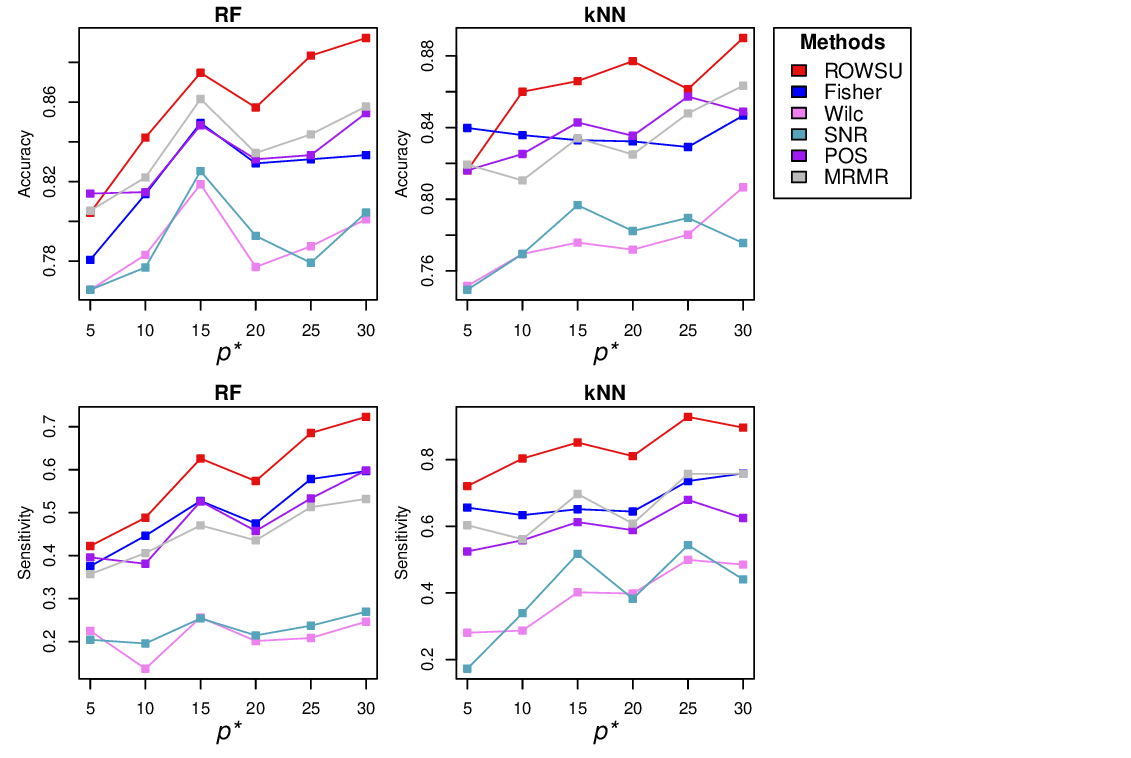}
	\caption{Matplots of the results computed for the top $p^*$ features by the proposed method and other state-of-the-art procedures using the \textbf{$D_2$} dataset.}
	\label{m2}
\end{figure}

\begin{figure}
	\centering
	\includegraphics[width=0.9\textwidth]{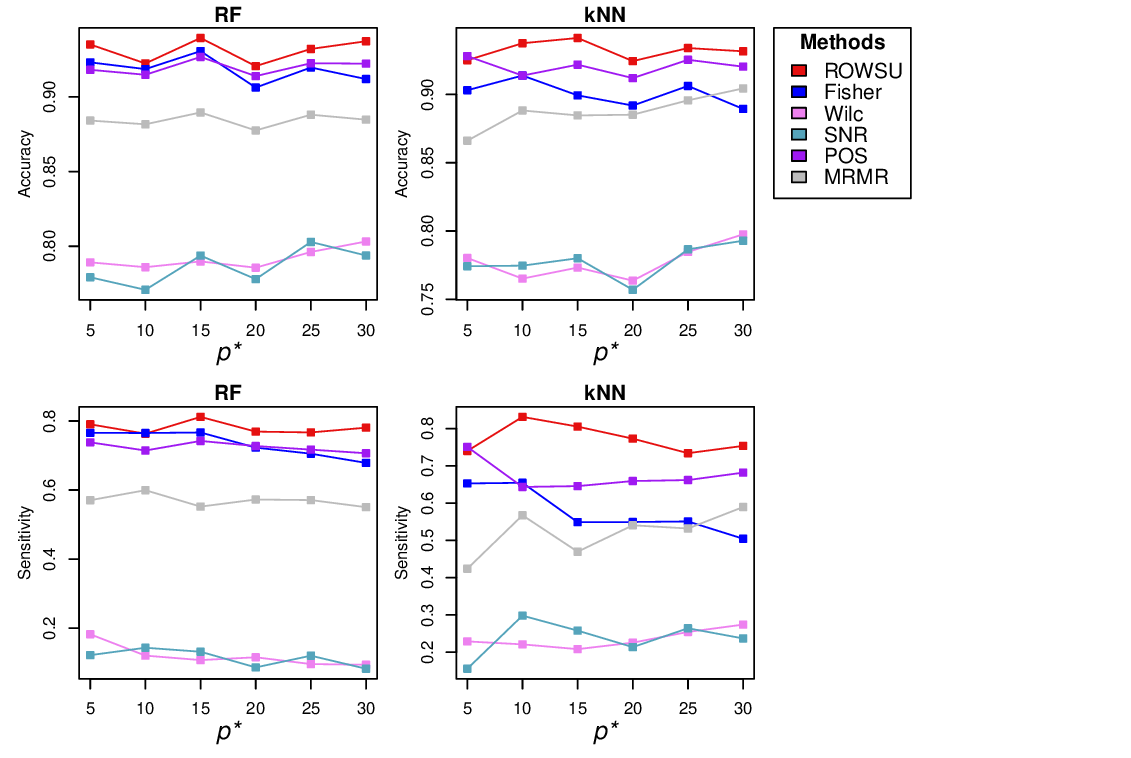}
	\caption{Matplots of the results computed for the top $p^*$ features by the proposed method and other state-of-the-art procedures using the \textbf{$D_3$} dataset.}
	\label{m3}
\end{figure}

\begin{figure}
	\centering
	\includegraphics[width=0.9\textwidth]{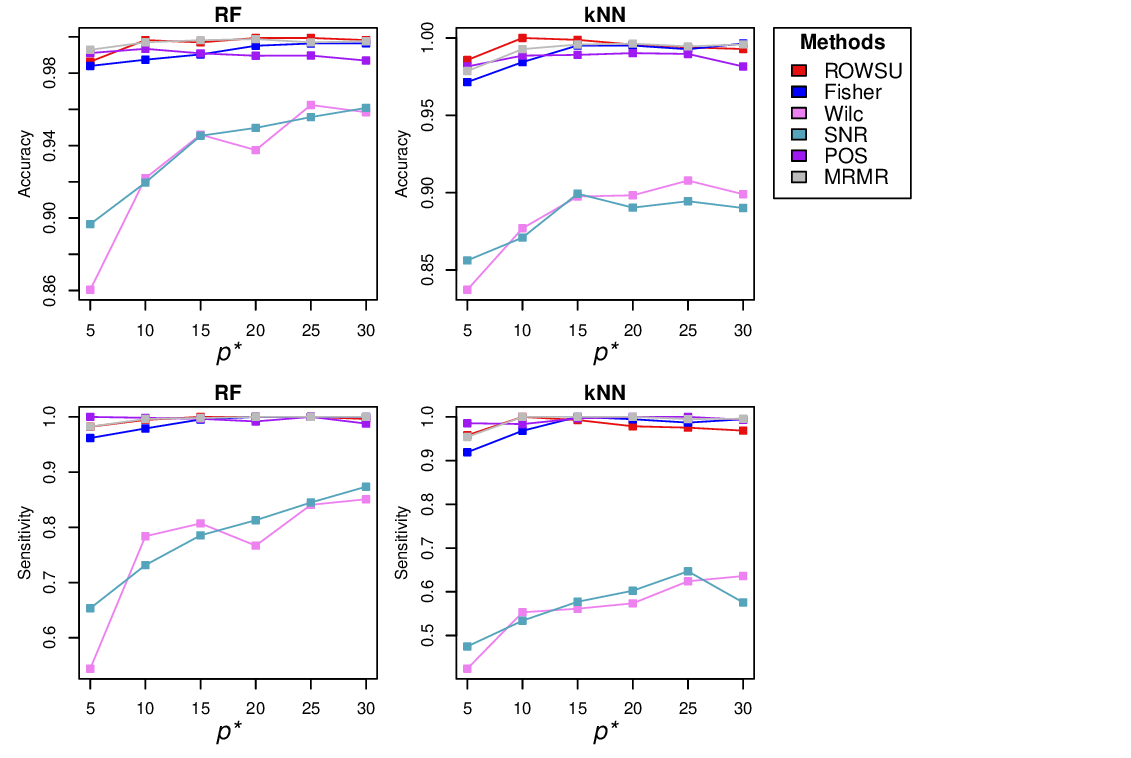}
	\caption{Matplots of the results computed for the top $p^*$ features by the proposed method and other state-of-the-art procedures using the \textbf{$D_4$} dataset.}
	\label{m4}
\end{figure}

\begin{figure}
	\centering
	\includegraphics[width=0.9\textwidth]{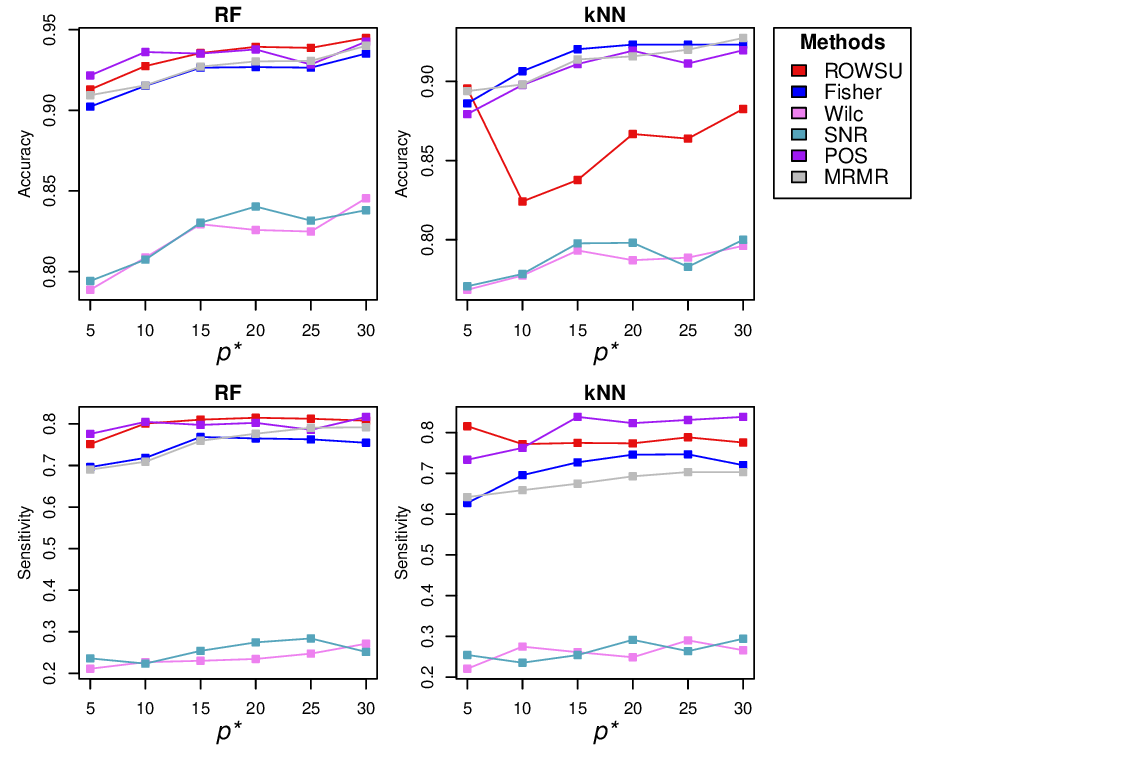}
	\caption{Matplots of the results computed for the top $p^*$ features by the proposed method and other state-of-the-art procedures using the \textbf{$D_5$} dataset.}
	\label{m5}
\end{figure}

\begin{figure}
	\centering
	\includegraphics[width=0.9\textwidth]{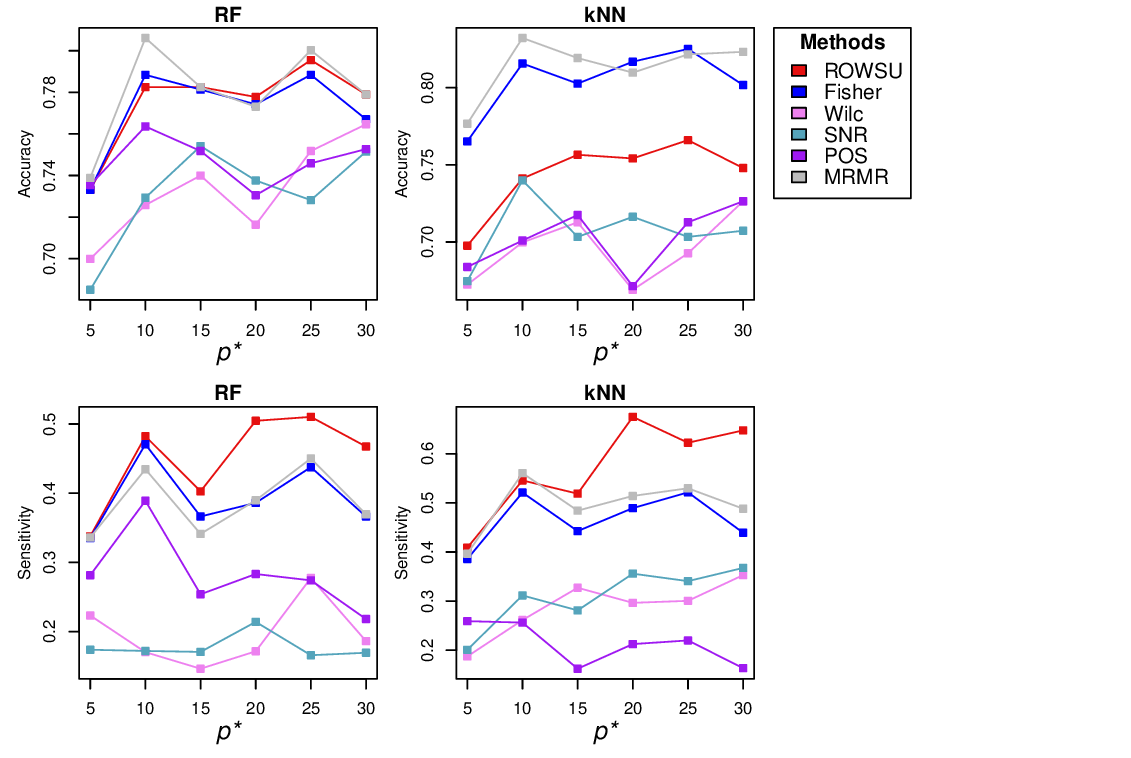}
	\caption{Matplots of the results computed for the top $p^*$ features by the proposed method and other state-of-the-art procedures using the \textbf{$D_6$} dataset.}
	\label{m6}
\end{figure}

\begin{figure}
	\centering
	\includegraphics[width=0.9\textwidth]{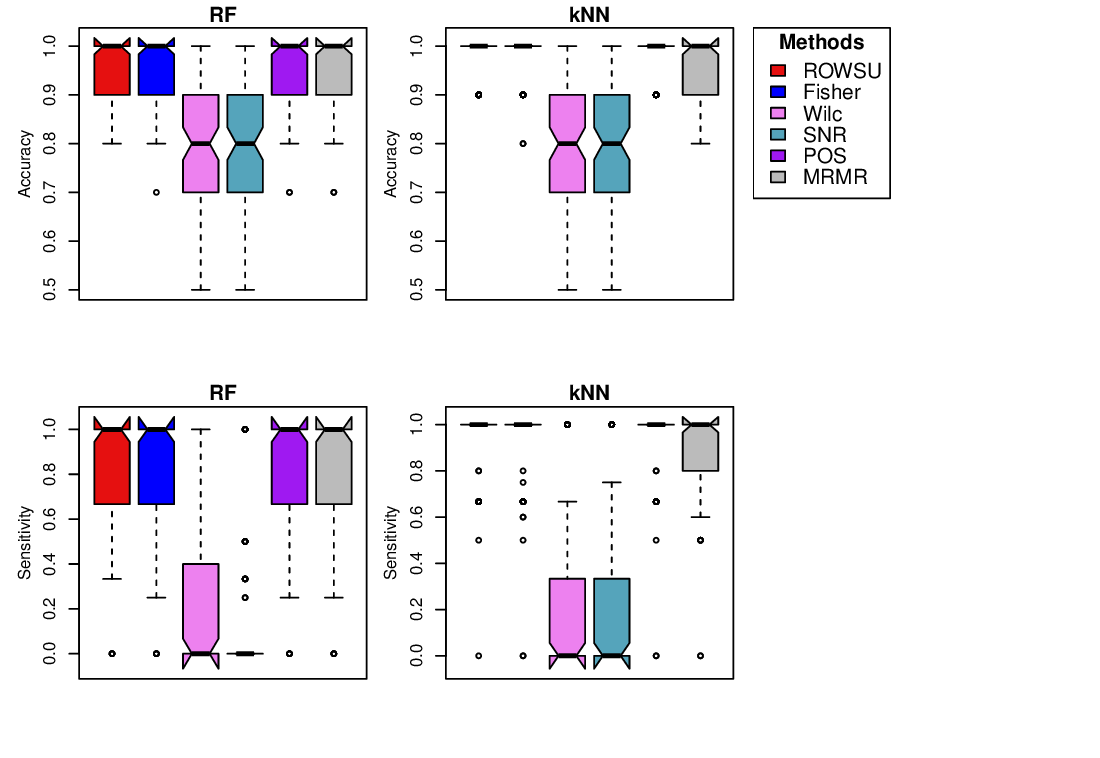}
	\caption{Boxplots of the results computed for the top 10 features by the proposed method and other state-of-the-art procedures using the \textbf{$D_1$} dataset.}
	\label{b1}
\end{figure}

\begin{figure}
	\centering
	\includegraphics[width=0.9\textwidth]{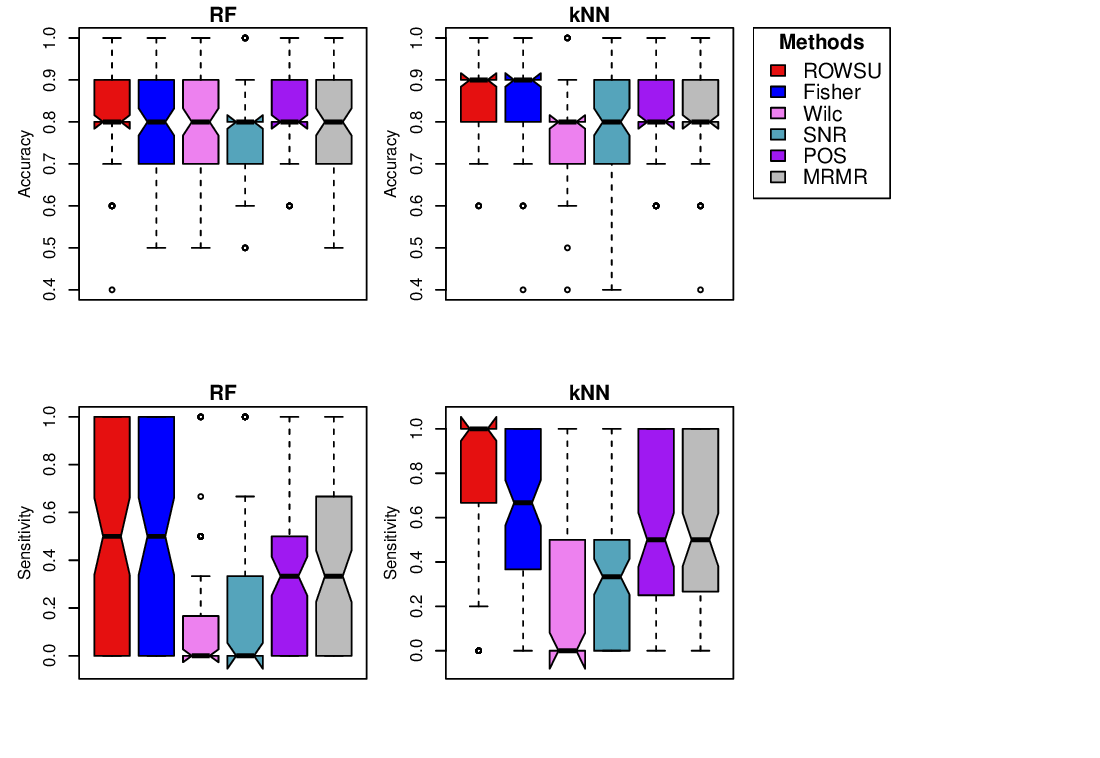}
	\caption{Boxplots of the results computed for the top 10 features by the proposed method and other state-of-the-art procedures using the \textbf{$D_2$} dataset.}
	\label{b2}
\end{figure}

\begin{figure}
	\centering
	\includegraphics[width=0.9\textwidth]{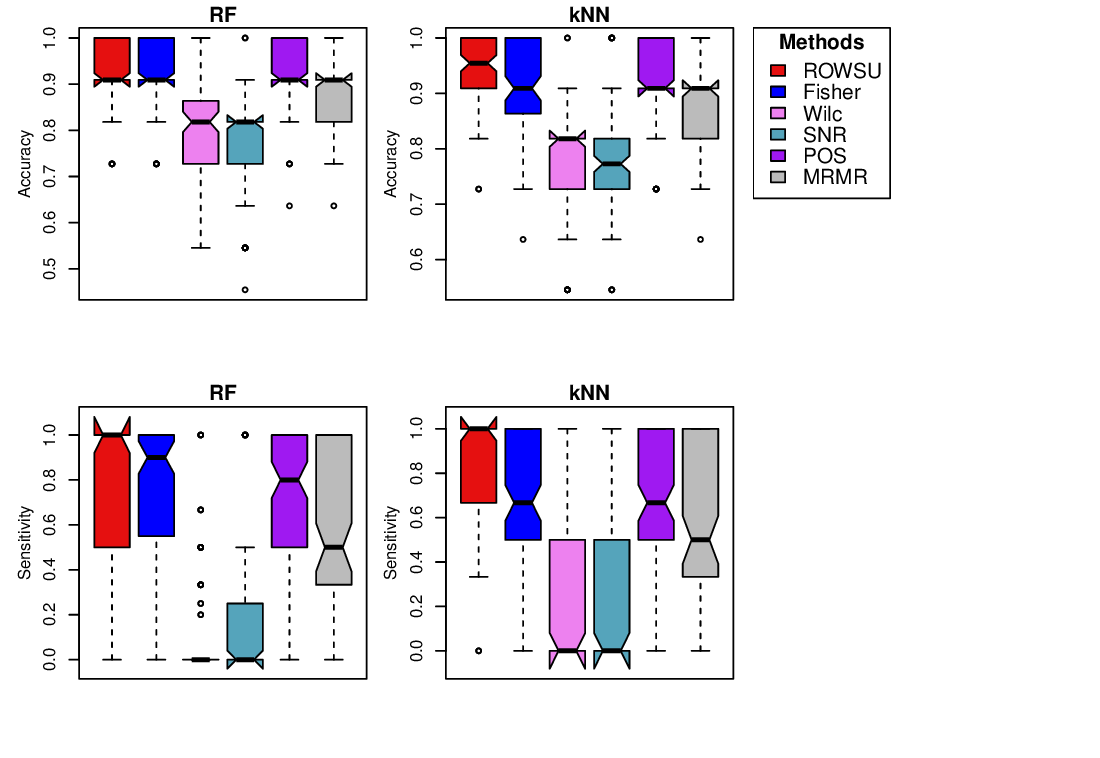}
	\caption{Boxplots of the results computed for the top 10 features by the proposed method and other state-of-the-art procedures using the \textbf{$D_3$} dataset.}
	\label{b3}
\end{figure}

\begin{figure}
	\centering
	\includegraphics[width=0.9\textwidth]{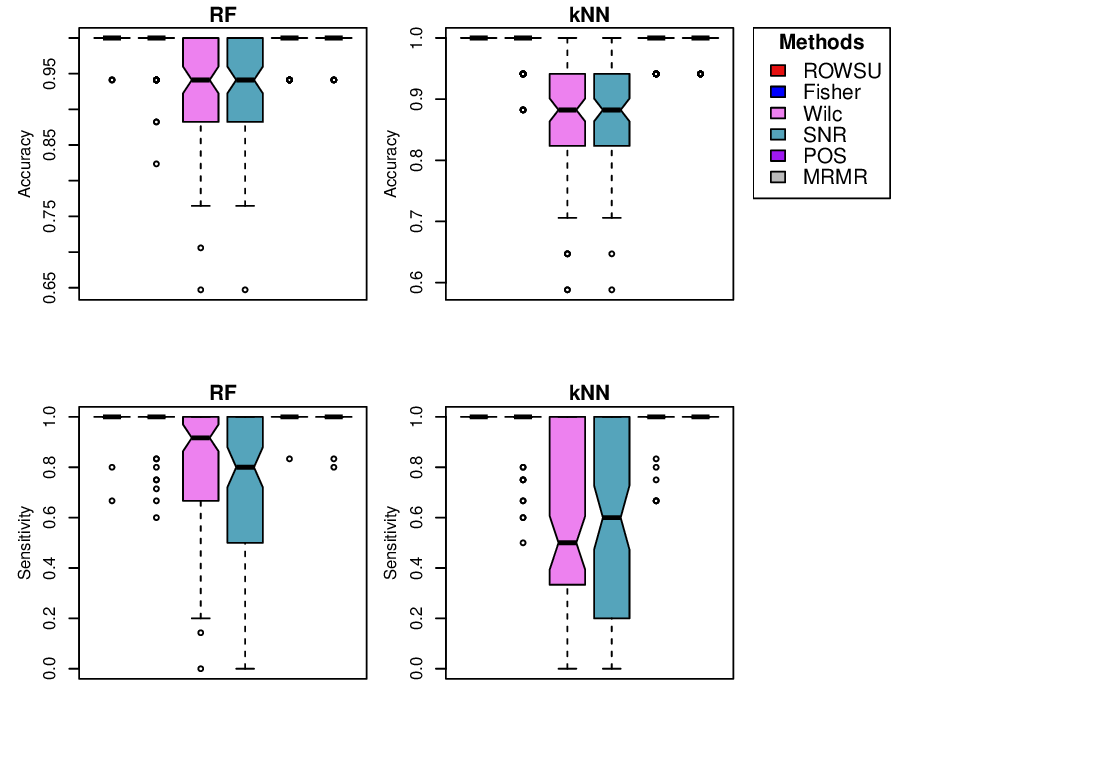}
	\caption{Boxplots of the results computed for the top 10 features by the proposed method and other state-of-the-art procedures using the \textbf{$D_4$} dataset.}
	\label{b4}
\end{figure}

\begin{figure}
    \centering
	\includegraphics[width=0.9\textwidth]{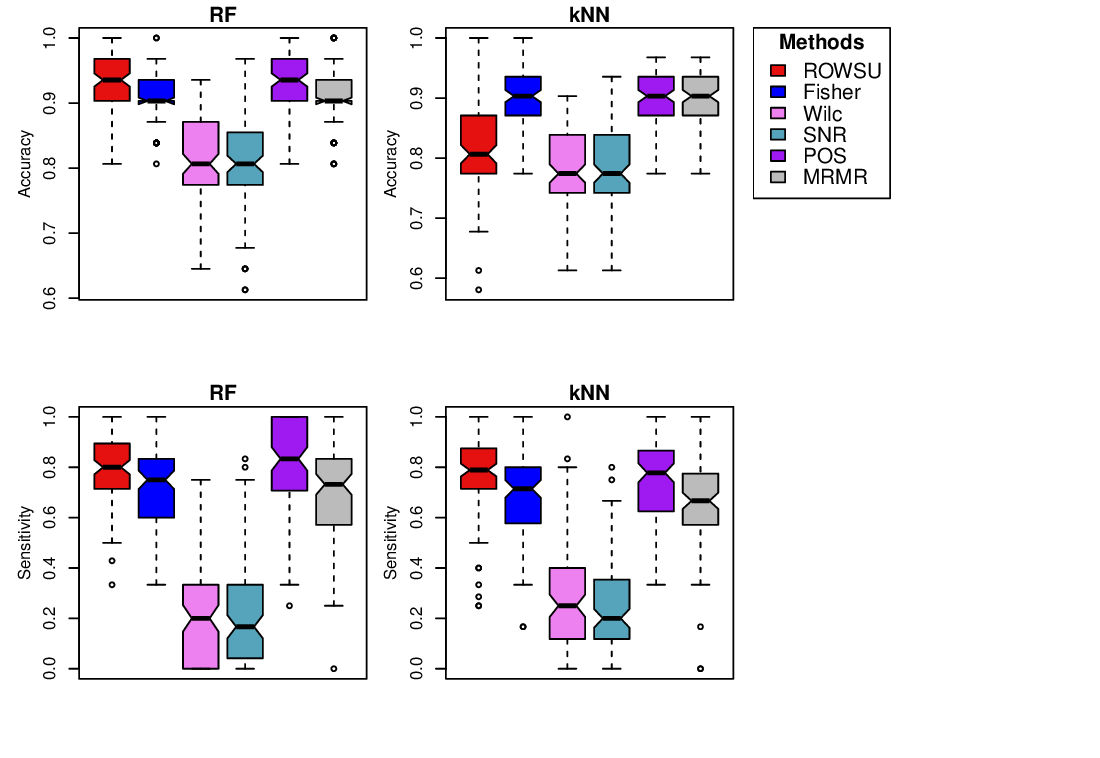}
	\caption{Boxplots of the results computed for the top 10 features by the proposed method and other state-of-the-art procedures using the \textbf{$D_5$} dataset.}
	\label{b5}
\end{figure}

\begin{figure}
	\centering
	\includegraphics[width=0.9\textwidth]{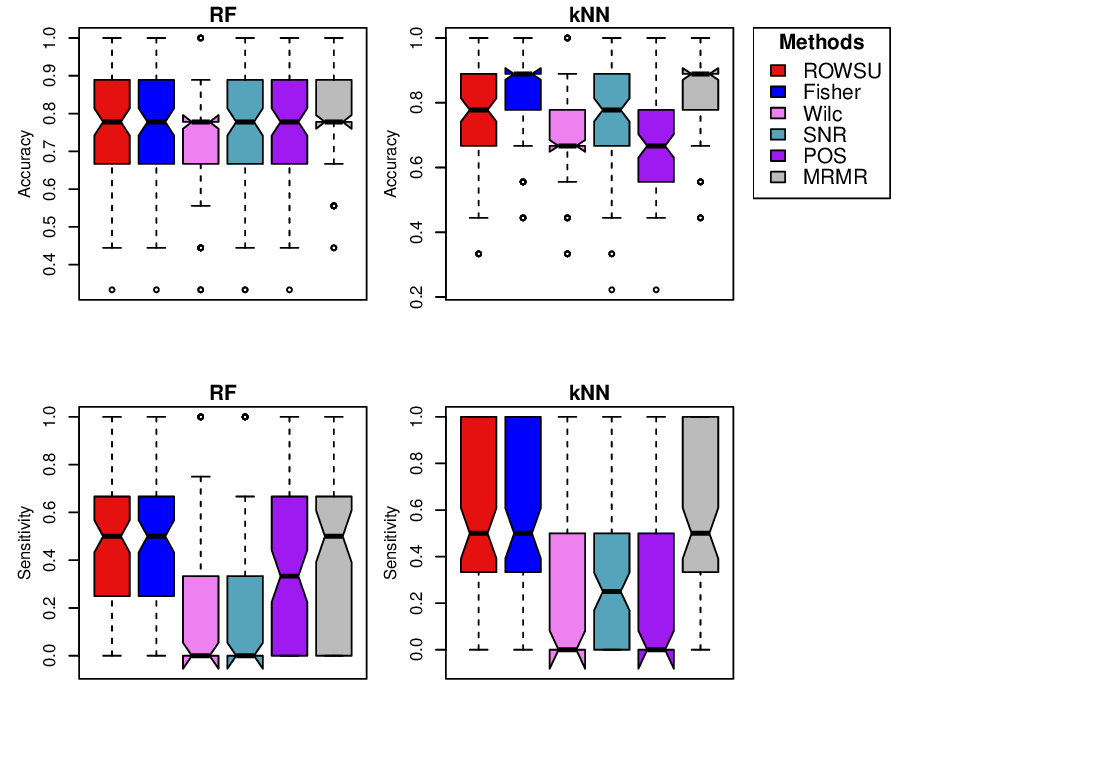}
	\caption{Boxplots of the results computed for the top 10 features by the proposed method and other state-of-the-art procedures using the \textbf{$D_6$} dataset.}
	\label{b6}
\end{figure}

Furthermore, similar conclusions can be drawn from the results based on $D_5$ and $D_6$ datasets. The results for these datasets are given in Tables \ref{d5} and \ref{d6}, respectively. For further evaluation, stability plots have also been constructed in Figures \ref{m1}-\ref{m6} and boxplots for the top 10 features selected by various procedures in Figures \ref{b1}-\ref{b6}. These figures demonstrate the consistency of the procedures, where the proposed method consistently performs better than the other methods in most of the cases.

\section{Conclusion}
\label{conclusion}
This work proposed a novel features selection method called robust weighted score for unbalanced data (ROWSU). It consists of a sequence of steps to achieve the desired results where existing methods perform poorly in the presence of a class-imbalanced problem. First, it balances the dataset using the observations in the minority class. Second, ROWSU identifies a minimum subset of the most discriminative features by using a greedy search approach in conjunction with a novel robust weighted score. The proposed method has given promising results when the class distribution of the dataset is extremely skewed. The proposed method is compared with state-of-the-art procedures on 6 benchmark datasets. The proposed ROWSU method outperformed the existing techniques in the majority of the cases. The performance of the ROWSU method and other feature selection methods are evaluated through the random forest (RF) and $k$ nearest neighbours ($k$NN) classifiers. Classification accuracy and sensitivity have been used as evaluation metrics. Moreover, stability plots of the results are constructed in the paper for different numbers of features which show the consistency of the proposed method. For a better understanding of the performance, boxplots of the results for the top 10 features have also been constructed. The results given in the paper demonstrated that the proposed method can effectively solve the class-imbalanced problem and discard non-informative and redundant genes. 

\begin{table}[h]
	\caption{Results calculated for the top-ranked $p^*$ feature selected by the proposed method and other state-of-the-art procedures using the \textbf{$D_1$} dataset.}
	\label{d1}
	\hspace{-0.7cm}
	\setlength{\tabcolsep}{2pt}
	\renewcommand{\arraystretch}{1}
	\begin{tabular}{cccccccccccccc}
		\toprule
		\multirow{2}{*}{Metric} & \multirow{2}{*}{\makecell{$p^*$}} & \multicolumn{6}{c}{RF} & \multicolumn{6}{c}{\textit{k}NN} \\   \cmidrule(lr){3-8} \cmidrule(lr){9-14}
		& 	&  ROWSU & Fish & Wilc & SNR & POS & MRMR &  ROWSU & Fish & Wilc & SNR & POS & MRMR \\
		\midrule
		\multirow{6}{*}{\begin{turn}{270}Accuracy\end{turn}} & 5 & \textbf{0.943} & \textbf{0.943} & 0.766 & 0.793 & 0.927 & 0.932 & \textbf{0.972} & 0.970 & 0.758 & 0.773 & 0.953 & 0.953 \\
		& 10 & \textbf{0.948} & 0.944 & 0.801 & 0.788 & \textbf{0.948} & 0.938 & 0.975 & 0.976 & 0.785 & 0.772 & \textbf{0.982} & 0.970 \\
		& 15 & \textbf{0.967} & 0.960 & 0.821 & 0.819 & 0.956 & 0.951 & 0.985 & \textbf{0.987} & 0.796 & 0.804 & 0.981 & 0.984 \\
		& 20 & \textbf{0.972} & 0.968 & 0.808 & 0.811 & 0.964 & 0.963 & \textbf{0.995} & 0.984 & 0.816 & 0.829 & 0.974 & 0.985 \\
		& 25 & \textbf{0.957} & 0.947 & 0.812 & 0.815 & 0.952 & 0.938 & \textbf{0.994} & 0.981 & 0.802 & 0.805 & 0.983 & 0.979 \\
		& 30 & \textbf{0.963} & 0.954 & 0.810 & 0.812 & 0.949 & 0.942 & \textbf{0.994} & 0.985 & 0.835 & 0.810 & 0.975 & 0.980 \\
		
		&  & & & & & & & & & & & & \\
		\multirow{6}{*}{\begin{turn}{270}Sensitivity\end{turn}} & 5 & \textbf{0.776} & 0.732 & 0.200 & 0.190 & 0.725 & 0.663 & \textbf{0.890} & 0.884 & 0.179 & 0.220 & 0.843 & 0.778 \\
		& 10 & \textbf{0.810} & 0.789 & 0.274 & 0.154 & 0.799 & 0.769 & \textbf{0.934} & 0.913 & 0.237 & 0.215 & 0.930 & 0.890 \\
		& 15 & \textbf{0.842} & 0.829 & 0.151 & 0.151 & 0.814 & 0.771 & \textbf{0.953} & 0.950 & 0.191 & 0.219 & 0.918 & 0.929 \\
		& 20 & \textbf{0.871} & \textbf{0.871} & 0.141 & 0.184 & 0.839 & 0.841 & \textbf{0.986} & 0.939 & 0.226 & 0.283 & 0.898 & 0.941 \\
		& 25 & \textbf{0.807} & 0.786 & 0.204 & 0.208 & 0.804 & 0.741 & \textbf{0.984} & 0.925 & 0.217 & 0.274 & 0.928 & 0.927 \\
		& 30 & \textbf{0.846} & 0.829 & 0.135 & 0.130 & 0.796 & 0.760 & \textbf{0.978} & 0.954 & 0.306 & 0.225 & 0.910 & 0.925 \\ \bottomrule
	\end{tabular}
\end{table}

\begin{table}[]
	\caption{Results calculated for the top-ranked $p^*$ feature selected by the proposed method and other state-of-the-art procedures using the \textbf{$D_2$} dataset.}
	\label{d2}
	\hspace{-0.7cm}
	\setlength{\tabcolsep}{2pt}
	\renewcommand{\arraystretch}{1}
	\begin{tabular}{cccccccccccccc}
		\toprule
		\multirow{2}{*}{Metric} & \multirow{2}{*}{\makecell{$p^*$}} & \multicolumn{6}{c}{RF} & \multicolumn{6}{c}{\textit{k}NN} \\   \cmidrule(lr){3-8} \cmidrule(lr){9-14}
		& 	&  ROWSU & Fish & Wilc & SNR & POS & MRMR &  ROWSU & Fish & Wilc & SNR & POS & MRMR \\
		\midrule
		\multirow{6}{*}{\begin{turn}{270}Accuracy\end{turn}} & 5 & 0.804 & 0.781 & 0.766 & 0.766 & 0.814 & \textbf{0.805} & 0.816 & \textbf{0.840} & 0.752 & 0.749 & 0.816 & 0.819 \\
		& 10 & \textbf{0.842} & 0.814 & 0.783 & 0.777 & 0.815 & 0.822 & \textbf{0.860} & 0.836 & 0.769 & 0.769 & 0.825 & 0.811 \\
		& 15 & \textbf{0.875} & 0.849 & 0.819 & 0.825 & 0.848 & 0.862 & \textbf{0.866} & 0.833 & 0.776 & 0.797 & 0.843 & 0.834 \\
		& 20 & \textbf{0.857} & 0.829 & 0.777 & 0.793 & 0.831 & 0.834 & \textbf{0.877} & 0.832 & 0.772 & 0.782 & 0.835 & 0.825 \\
		& 25 & \textbf{0.883} & 0.831 & 0.788 & 0.779 & 0.833 & 0.844 & \textbf{0.861} & 0.829 & 0.780 & 0.790 & 0.857 & 0.848 \\
		& 30 & \textbf{0.892} & 0.833 & 0.801 & 0.804 & 0.854 & 0.858 & \textbf{0.890} & 0.847 & 0.807 & 0.776 & 0.849 & 0.863 \\
		
		&  & & & & & & & & & & & & \\
		\multirow{6}{*}{\begin{turn}{270}Sensitivity\end{turn}} & 5 & \textbf{0.422} & 0.375 & 0.225 & 0.204 & 0.396 & 0.357 & \textbf{0.721} & 0.656 & 0.280 & 0.172 & 0.524 & 0.603 \\
		& 10 & \textbf{0.488} & 0.446 & 0.137 & 0.196 & 0.381 & 0.406 & \textbf{0.804} & 0.634 & 0.287 & 0.339 & 0.558 & 0.561 \\
		& 15 & \textbf{0.626} & 0.527 & 0.256 & 0.254 & 0.526 & 0.471 & \textbf{0.851} & 0.651 & 0.402 & 0.517 & 0.613 & 0.697 \\
		& 20 & \textbf{0.574} & 0.475 & 0.201 & 0.214 & 0.457 & 0.436 & \textbf{0.811} & 0.644 & 0.398 & 0.382 & 0.588 & 0.608 \\
		& 25 & \textbf{0.685} & 0.578 & 0.208 & 0.237 & 0.533 & 0.513 & \textbf{0.928} & 0.736 & 0.499 & 0.543 & 0.679 & 0.757 \\
		& 30 & \textbf{0.723} & 0.597 & 0.246 & 0.269 & 0.599 & 0.532 & \textbf{0.896} & 0.759 & 0.485 & 0.441 & 0.625 & 0.758 \\ \bottomrule
	\end{tabular}
\end{table}

\begin{table}[]
	\caption{Results calculated for the top-ranked $p^*$ feature selected by the proposed method and other state-of-the-art procedures using the \textbf{$D_3$} dataset.}
	\label{d3}
	\hspace{-0.7cm}
	\setlength{\tabcolsep}{2pt}
	\renewcommand{\arraystretch}{1}
	\begin{tabular}{cccccccccccccc}
		\toprule
		\multirow{2}{*}{Metric} & \multirow{2}{*}{\makecell{$p^*$}} & \multicolumn{6}{c}{RF} & \multicolumn{6}{c}{\textit{k}NN} \\   \cmidrule(lr){3-8} \cmidrule(lr){9-14}
		& 	&  ROWSU & Fish & Wilc & SNR & POS & MRMR &  ROWSU & Fish & Wilc & SNR & POS & MRMR \\
		\midrule
		\multirow{6}{*}{\begin{turn}{270}Accuracy\end{turn}} & 5 & \textbf{0.935} & 0.923 & 0.789 & 0.779 & 0.918 & 0.884 & 0.925 & 0.903 & 0.780 & 0.774 & \textbf{0.928} & 0.866 \\
		& 10 & \textbf{0.922} & 0.919 & 0.786 & 0.771 & 0.915 & 0.882 & \textbf{0.938} & 0.914 & 0.765 & 0.775 & 0.914 & 0.888 \\
		& 15 & \textbf{0.939} & 0.931 & 0.790 & 0.794 & 0.927 & 0.890 & \textbf{0.941} & 0.899 & 0.773 & 0.780 & 0.922 & 0.885 \\
		& 20 & \textbf{0.921} & 0.906 & 0.786 & 0.778 & 0.914 & 0.878 & \textbf{0.924} & 0.892 & 0.764 & 0.757 & 0.912 & 0.885 \\
		& 25 & \textbf{0.932} & 0.920 & 0.796 & 0.803 & 0.922 & 0.888 & \textbf{0.934} & 0.906 & 0.785 & 0.787 & 0.925 & 0.896 \\
		& 30 & \textbf{0.937} & 0.912 & 0.803 & 0.794 & 0.922 & 0.885 & \textbf{0.932} & 0.889 & 0.798 & 0.793 & 0.920 & 0.904 \\
		
		&  & & & & & & & & & & & & \\
		\multirow{6}{*}{\begin{turn}{270}Sensitivity\end{turn}} & 5 & \textbf{0.790} & 0.766 & 0.182 & 0.122 & 0.738 & 0.571 & 0.740 & 0.653 & 0.229 & 0.156 & \textbf{0.751} & 0.424 \\
		& 10 & 0.763 & \textbf{0.765} & 0.120 & 0.143 & 0.714 & 0.599 & \textbf{0.831} & 0.655 & 0.221 & 0.298 & 0.643 & 0.567 \\
		& 15 & \textbf{0.812} & 0.767 & 0.108 & 0.132 & 0.742 & 0.552 & \textbf{0.805} & 0.549 & 0.208 & 0.258 & 0.646 & 0.470 \\
		& 20 & \textbf{0.769} & 0.723 & 0.116 & 0.086 & 0.727 & 0.572 & \textbf{0.773} & 0.549 & 0.226 & 0.214 & 0.659 & 0.541 \\
		& 25 & \textbf{0.767} & 0.705 & 0.096 & 0.120 & 0.717 & 0.571 & \textbf{0.734} & 0.551 & 0.254 & 0.264 & 0.662 & 0.532 \\
		& 30 & \textbf{0.781} & 0.679 & 0.094 & 0.082 & 0.707 & 0.550 & \textbf{0.754} & 0.504 & 0.274 & 0.237 & 0.682 & 0.590 \\ \bottomrule
	\end{tabular}
\end{table}

\begin{table}[]
	\caption{Results calculated for the top-ranked $p^*$ feature selected by the proposed method and other state-of-the-art procedures using the \textbf{$D_4$} dataset.}
	\label{d4}
	\hspace{-0.7cm}
	\setlength{\tabcolsep}{2pt}
	\renewcommand{\arraystretch}{1}
	\begin{tabular}{cccccccccccccc}
		\toprule
		\multirow{2}{*}{Metric} & \multirow{2}{*}{\makecell{$p^*$}} & \multicolumn{6}{c}{RF} & \multicolumn{6}{c}{\textit{k}NN} \\   \cmidrule(lr){3-8} \cmidrule(lr){9-14}
		& 	&  ROWSU & Fish & Wilc & SNR & POS & MRMR &  ROWSU & Fish & Wilc & SNR & POS & MRMR \\
		\midrule
		\multirow{6}{*}{\begin{turn}{270}Accuracy\end{turn}} & 5 & 0.986 & 0.984 & 0.860 & 0.897 & 0.991 & \textbf{0.993} & \textbf{0.986} & 0.971 & 0.837 & 0.856 & 0.982 & 0.979 \\
		& 10 & \textbf{0.998} & 0.987 & 0.922 & 0.920 & 0.993 & 0.997 & \textbf{1.000} & 0.984 & 0.877 & 0.871 & 0.989 & 0.993 \\
		& 15 & 0.997 & 0.990 & 0.946 & 0.945 & 0.991 & \textbf{0.998} & \textbf{0.999} & 0.995 & 0.898 & 0.899 & 0.989 & 0.996 \\
		& 20 & \textbf{0.999} & 0.995 & 0.938 & 0.950 & 0.990 & \textbf{0.999} & \textbf{0.996} & 0.995 & 0.898 & 0.890 & 0.990 & \textbf{0.996} \\
		& 25 & \textbf{0.999} & 0.996 & 0.962 & 0.956 & 0.990 & 0.997 & 0.994 & 0.993 & 0.908 & 0.894 & 0.990 & \textbf{0.995} \\
		& 30 & \textbf{0.998} & 0.996 & 0.958 & 0.961 & 0.987 & \textbf{0.998} & 0.993 & \textbf{0.996} & 0.899 & 0.890 & 0.982 & \textbf{0.996} \\
		
		&  & & & & & & & & & & & & \\
		\multirow{6}{*}{\begin{turn}{270}Sensitivity\end{turn}} & 5 & 0.982 & 0.962 & 0.544 & 0.653 & \textbf{1.000} & 0.982 & 0.958 & 0.919 & 0.424 & 0.475 & \textbf{0.986} & 0.954 \\
		& 10 & 0.995 & 0.979 & 0.784 & 0.732 & \textbf{0.998} & 0.996 & \textbf{1.000} & 0.968 & 0.553 & 0.534 & 0.984 & \textbf{1.000} \\
		& 15 & \textbf{1.000} & 0.995 & 0.807 & 0.785 & 0.996 & 0.998 & 0.993 & \textbf{1.000} & 0.561 & 0.577 & 0.997 & \textbf{1.000} \\
		& 20 & \textbf{1.000} & \textbf{1.000} & 0.767 & 0.813 & 0.992 & \textbf{1.000} & 0.978 & 0.995 & 0.574 & 0.602 & \textbf{1.000} & \textbf{1.000} \\
		& 25 & \textbf{1.000} & \textbf{1.000} & 0.841 & 0.845 & \textbf{1.000} & \textbf{1.000} & 0.975 & 0.987 & 0.624 & 0.647 & \textbf{1.000} & 0.995 \\
		& 30 & 0.996 & \textbf{1.000} & 0.851 & 0.874 & 0.988 & \textbf{1.000} & 0.969 & 0.995 & 0.636 & 0.576 & 0.994 & \textbf{0.996} \\ \bottomrule
	\end{tabular}
\end{table}

\begin{table}[]
	\caption{Results calculated for the top-ranked $p^*$ feature selected by the proposed method and other state-of-the-art procedures using the \textbf{$D_5$} dataset.}
	\label{d5}
	\hspace{-0.7cm}
	\setlength{\tabcolsep}{2pt}
	\renewcommand{\arraystretch}{1}
	\begin{tabular}{cccccccccccccc}
		\toprule
		\multirow{2}{*}{Metric} & \multirow{2}{*}{\makecell{$p^*$}} & \multicolumn{6}{c}{RF} & \multicolumn{6}{c}{\textit{k}NN} \\   \cmidrule(lr){3-8} \cmidrule(lr){9-14}
		& 	&  ROWSU & Fish & Wilc & SNR & POS & MRMR &  ROWSU & Fish & Wilc & SNR & POS & MRMR \\
		\midrule
		\multirow{6}{*}{\begin{turn}{270}Accuracy\end{turn}} & 5 & 0.913 & 0.902 & 0.789 & 0.794 & \textbf{0.922} & 0.909 & \textbf{0.895} & 0.886 & 0.768 & 0.771 & 0.879 & 0.894 \\
		& 10 & 0.927 & 0.915 & 0.809 & 0.807 & \textbf{0.936} & 0.915 & 0.824 & \textbf{0.906} & 0.777 & 0.778 & 0.898 & 0.898 \\
		& 15 & \textbf{0.935} & 0.926 & 0.829 & 0.830 & \textbf{0.935} & 0.927 & 0.838 & \textbf{0.920} & 0.793 & 0.798 & 0.911 & 0.914 \\
		& 20 & \textbf{0.939} & 0.927 & 0.826 & 0.840 & 0.938 & 0.930 & 0.867 & \textbf{0.923} & 0.787 & 0.798 & 0.919 & 0.916 \\
		& 25 & \textbf{0.939} & 0.926 & 0.825 & 0.832 & 0.928 & 0.931 & 0.864 & \textbf{0.923} & 0.789 & 0.783 & 0.911 & 0.920 \\
		& 30 & \textbf{0.945} & 0.935 & 0.845 & 0.838 & 0.943 & 0.940 & 0.883 & 0.923 & 0.796 & 0.800 & 0.920 & \textbf{0.927} \\
		
		&  & & & & & & & & & & & & \\
	    \multirow{6}{*}{\begin{turn}{270}Sensitivity\end{turn}} & 5 & 0.751 & 0.696 & 0.211 & 0.236 & \textbf{0.776} & 0.690 & \textbf{0.816} & 0.627 & 0.221 & 0.254 & 0.733 & 0.642 \\
		& 10 & 0.801 & 0.718 & 0.227 & 0.223 & \textbf{0.805} & 0.709 & \textbf{0.772} & 0.696 & 0.275 & 0.235 & 0.762 & 0.659 \\
		& 15 & \textbf{0.810} & 0.768 & 0.230 & 0.254 & 0.798 & 0.759 & 0.775 & 0.727 & 0.261 & 0.254 & \textbf{0.838} & 0.675 \\
		& 20 & \textbf{0.815} & 0.765 & 0.235 & 0.274 & 0.803 & 0.776 & 0.774 & 0.746 & 0.249 & 0.292 & \textbf{0.823} & 0.693 \\
		& 25 & \textbf{0.812} & 0.763 & 0.247 & 0.284 & 0.786 & 0.791 & 0.788 & 0.747 & 0.290 & 0.264 & \textbf{0.831} & 0.703 \\
		& 30 & 0.808 & 0.755 & 0.271 & 0.252 & \textbf{0.817} & 0.792 & 0.776 & 0.720 & 0.266 & 0.294 & \textbf{0.838} & 0.703 \\ \bottomrule
	\end{tabular}
\end{table}

\begin{table}[]
	\caption{Results calculated for the top-ranked $p^*$ feature selected by the proposed method and other state-of-the-art procedures using the \textbf{$D_6$} dataset.}
	\label{d6}
	\hspace{-0.7cm}
	\setlength{\tabcolsep}{2pt}
	\renewcommand{\arraystretch}{1}
	\begin{tabular}{cccccccccccccc}
		\toprule
		\multirow{2}{*}{Metric} & \multirow{2}{*}{\makecell{$p^*$}} & \multicolumn{6}{c}{RF} & \multicolumn{6}{c}{\textit{k}NN} \\   \cmidrule(lr){3-8} \cmidrule(lr){9-14}
		& 	&  ROWSU & Fish & Wilc & SNR & POS & MRMR &  ROWSU & Fish & Wilc & SNR & POS & MRMR \\
		\midrule
		\multirow{6}{*}{\begin{turn}{270}Accuracy\end{turn}} & 5 & 0.733 & 0.733 & 0.700 & 0.685 & 0.735 & \textbf{0.739} & 0.698 & 0.765 & 0.672 & 0.675 & 0.684 & \textbf{0.777} \\
		& 10 & 0.783 & 0.788 & 0.726 & 0.729 & 0.764 & \textbf{0.806} & 0.741 & 0.816 & 0.700 & 0.740 & 0.701 & \textbf{0.832} \\
		& 15 & \textbf{0.783} & 0.781 & 0.740 & 0.754 & 0.752 & \textbf{0.783} & 0.757 & 0.803 & 0.713 & 0.703 & 0.717 & 0.819 \\
		& 20 & 0.778 & 0.774 & 0.716 & 0.738 & 0.730 & 0.773 & 0.754 & \textbf{0.817} & 0.669 & 0.716 & 0.671 & 0.810 \\
		& 25 & 0.796 & 0.788 & 0.752 & 0.728 & 0.746 & \textbf{0.800} & 0.766 & \textbf{0.825} & 0.693 & 0.703 & 0.713 & 0.822 \\
		& 30 & \textbf{0.779} & 0.767 & 0.765 & 0.751 & 0.753 & \textbf{0.779} & 0.748 & 0.802 & 0.726 & 0.707 & 0.726 & 0.823 \\
		
		&  & & & & & & & & & & & & \\
		\multirow{6}{*}{\begin{turn}{270}Sensitivity\end{turn}} & 5 & 0.338 & 0.335 & 0.223 & 0.174 & 0.281 & 0.336 & \textbf{0.409} & 0.386 & 0.187 & 0.201 & 0.259 & 0.396 \\
		& 10 & \textbf{0.482} & 0.471 & 0.170 & 0.172 & 0.389 & 0.435 & 0.546 & 0.521 & 0.261 & 0.312 & 0.256 & \textbf{0.561} \\
		& 15 & \textbf{0.403} & 0.366 & 0.146 & 0.171 & 0.254 & 0.341 & \textbf{0.519} & 0.443 & 0.327 & 0.281 & 0.162 & 0.484 \\
		& 20 & \textbf{0.505} & 0.386 & 0.171 & 0.214 & 0.283 & 0.390 & \textbf{0.675} & 0.489 & 0.296 & 0.356 & 0.212 & 0.514 \\
		& 25 & \textbf{0.510} & 0.438 & 0.277 & 0.166 & 0.274 & 0.450 & \textbf{0.623} & 0.521 & 0.301 & 0.341 & 0.220 & 0.530 \\
		& 30 & \textbf{0.467} & 0.366 & 0.186 & 0.169 & 0.218 & 0.370 & \textbf{0.648} & 0.439 & 0.353 & 0.368 & 0.163 & 0.488 \\ \bottomrule
	\end{tabular}
\end{table}



\onecolumn
\section*{References}
\bibliographystyle{unsrt}
\bibliography{WileySTAT}
\end{document}